\documentclass[12pt]{article}

\usepackage[utf8]{inputenc}
\usepackage{amsmath, amssymb, amsthm, mathtools, bbm, graphicx, fullpage, caption, subcaption, multirow, url, authblk, hyperref}
\usepackage[dvipsnames]{xcolor}
\usepackage[normalem]{ulem}
\usepackage{setspace}
\usepackage{enumitem}

\usepackage{acro}
\DeclareAcronym{is}{short=IS, long=Interdaily Stability}
\DeclareAcronym{iv}{short=IV, long=Intradaily Variability}
\DeclareAcronym{m10}{short=IS, long=Most Active 10 Hours}
\DeclareAcronym{l5}{short=IV, long=Least Active 5 Hours}
\DeclareAcronym{ra}{short=RA, long=Relative Amplitude}

\title{Benchmarking changepoint detection algorithms on cardiac time series}

\author[1,2]{\small{Ayse Selen Cakmak}}
\author[1,3]{\small{Erik Reinertsen}}
\author[1,4]{\small{Shamim Nemati}} 
\author[1,3]{\small{Gari~D.~Clifford}}

\affil[1]{\small{Department of Biomedical Informatics, Emory University}}
\affil[2]{\small{School of Electrical and Computer Engineering, Georgia Institute of Technology}}
\affil[3]{\small{Department of Biomedical Informatics, Emory University}}
\affil[4]{\small{Department of Biomedical Informatics, University of California at San Diego}}

\begin{document}

\maketitle

\begin{abstract}
The pattern of state changes in a biomedical time series can be related to health or disease. This work presents a principled approach for selecting a changepoint detection algorithm for a specific task, such as disease classification. Eight key algorithms were compared, and the performance of each algorithm was evaluated as a function of temporal tolerance, noise, and abnormal conduction (ectopy) on realistic artificial cardiovascular time series data. All algorithms were applied to real data (cardiac time series of 22 patients with REM-behavior disorder (RBD) and 15 healthy controls) using the parameters selected on artificial data. Finally, features were derived from the detected changepoints to classify RBD patients from healthy controls using a K-Nearest Neighbors approach. On artificial data, Modified Bayesian Changepoint Detection algorithm provided superior positive predictive value for state change identification while Recursive Mean Difference Maximization (RMDM) achieved the highest true positive rate. For the classification task, features derived from the RMDM algorithm provided the highest leave one out cross validated accuracy of 0.89 and true positive rate of 0.87. Automatically detected changepoints provide useful information about subject's physiological state which cannot be directly observed. However, the choice of change point detection algorithm depends on the nature of the underlying data and the downstream application, such as a classification task. This work represents the first time change point detection algorithms have been compared in a meaningful way and utilized in a classification task, which demonstrates the effect of changepoint algorithm choice on application performance.


\end{abstract}

\section{Introduction}
\label{sec:introduction}

Non-stationarity is one of the key characteristics of human physiology and activity, driven by structured working days, alarms, unpredictable human interaction, etc., as well as intrinsic changes in the central nervous and cardiovascular systems. Many techniques have been proposed to artificially remove non-stationarities such as `detrending', or removing a mean, slope, or nonlinear fit in an arbitrary piecewise manner \cite{Lan2012, Wu2007}. However, such approaches tend to create large artifacts around changes in stationarity \cite{RAFFALOVICH1994, Nelson1981}. Moreover, many useful time series analysis techniques assume stationarity. Therefore, changepoint detection (CPD) -- the estimation of points in time where the probability distribution of a stochastic process changes -- can enable the analysis of stationary segments of data and reveal underlying structure. For instance, Bernaola-Galv\`an \cite{Bernaola-Galvan2001} used time series segmentation and CPD to investigate non-stationaries in human HR time series and found mean level jumps between HR segments were smaller in heart failure patients compared to healthy controls. Furthermore, HR interval segments were found to follow a power law distribution, for both heart failure patients and healthy controls.

In this work, eight CPD methods were applied to a realistic artificial dataset, which consisted of beat-to-beat (RR) interval time series data generated by a previously published model \cite{McSharry2002}, hereafter referred to as ``RRGen''. The performance of each CPD algorithm was evaluated as a function of noise, tolerance (time between a true and estimated changepoint), and arrhythmia (ectopy). Using artificial data enables the assessment of algorithmic performance by comparing estimated changepoints against true changepoints. Moreover, if the artificial dataset is realistic, parameters of the changepoint detection method can be optimized for use on real data that exhibits similar statistics. By using the knowledge of the exact time of state transitions inherent in RRGen, we tuned parameters of each algorithm to detect changepoints during in real overnight recordings. Theoretical results were extended to a real-world classification application in order to measure of the utility of changepoint detection techniques and to assess the applicability of tuning parameters of each algorithm using artificial data. Specifically, we extracted parameters of the power-law behavior in the distribution of segment lengths (time between each changepoint) to create a novel set of features for classifying individuals with REM Behavior Disorder (RBD).

RR intervals and sleep stages were investigated from 22 patients with RDB and 15 healthy controls. During sleep, the transition into the REM state is followed by tonic and phasic changes which include muscle atonia with bursts of cardiorespiratory variability \cite{lapierre1992polysomnographic}. However, patients with RBD lose muscle atonia and act out their dreams, which disrupts sleep continuity \cite{schenck2002rem}. In previous studies, heart rate variability metrics of patients with RBD were compared to controls \cite{postuma2010cardiac, ferini1996cardiac, lanfranchi2007cardiac}. Moreover, number of epochs classified as wake using actigraphy was used as a feature in building classifier between normal and RBD groups \cite{louter2014actigraphy}. This classifier resulted in a specificity of 0.96, but at with a very low sensitivity of 0.20 (which is equivalent to an accuracy of 0.58 on a balanced data set).

RBD patients have altered regulation of sleep-wake stages due to neuroanatomical changes in their brain \cite{boucetta2016structural}. Along with movement artifacts occurring more frequently due to nature of RBD, these alterations can affect changepoints detected from cardiac time series. In our approach, the length of time between changepoints was fit to parametric distributions, and the parameters of the model were used to classify patient groups. In this way, predictive power was tested, rather than simple statistical differences, with the understanding that a given level of significance does not necessarily translate into a useful metric \cite{Gelman2014,Lo2015,Gelman2017}. An optimal changepoint detection method which follows an underlying generative process would, in theory, be responsive to changes in the physiological conditions. Therefore, this method may result in a meaningful separation between patients and healthy controls and provide the high predictive power. (We note that because heart rate changes are not specific to sleep stages, and because sleep stages are scored on a 30 second epoch grid, they rarely correspond to real physiological changes.)

\section{Materials and methods}
Changepoint detection algorithms were assessed on a synthetic dataset, created using a realistic physiological model of heart rate (\cite{McSharry2002}) using the open source implementation `RRGen'. Using this model synthetic 24-hour data were generated 400 times with different random seeds, and the methods described below were applied to evaluate performance. Next, these methods were applied to the Physionet Cyclic Alternating Pattern Database consisting of polysomnographic recordings and Electrocardiography (ECG) traces that include subjects with RBD \cite{terzano2002atlas,PhysioNet} by using the optimal parameters derived in this step. The objective was to see if the statistics of the distribution of the times between state changes provided enough information to classify RDB patients from controls. 

\subsection{Artificial data}
The `RRGen' algorithm was used to generate realistic 24-hour RR time series (tachograms) from a model of cardiovascular interactions and transitions between physiological states \cite{McSharry2002}. The model incorporates short-range variability due to Meyer waves and respiratory sinus arrhythmia, and long-range transitions in physiological states, by using switching distributions extracted from real data. The model incorporates both short and long-term variability, modeled on the normal sinus rhythm database \cite{goldberger2000}. Different tachograms can be produced by calling RRGen with different seeds. Realistic ectopy and artifact were also added at incremental levels in a realistic manner (with a probability per beat ranging from 0\% to 1.2\%) per Clifford and McSharry \cite{clifford2002characterizing}.

\begin{figure}[ht]
\vspace{-4.0mm}
\centering
    \includegraphics[width=0.95\linewidth]{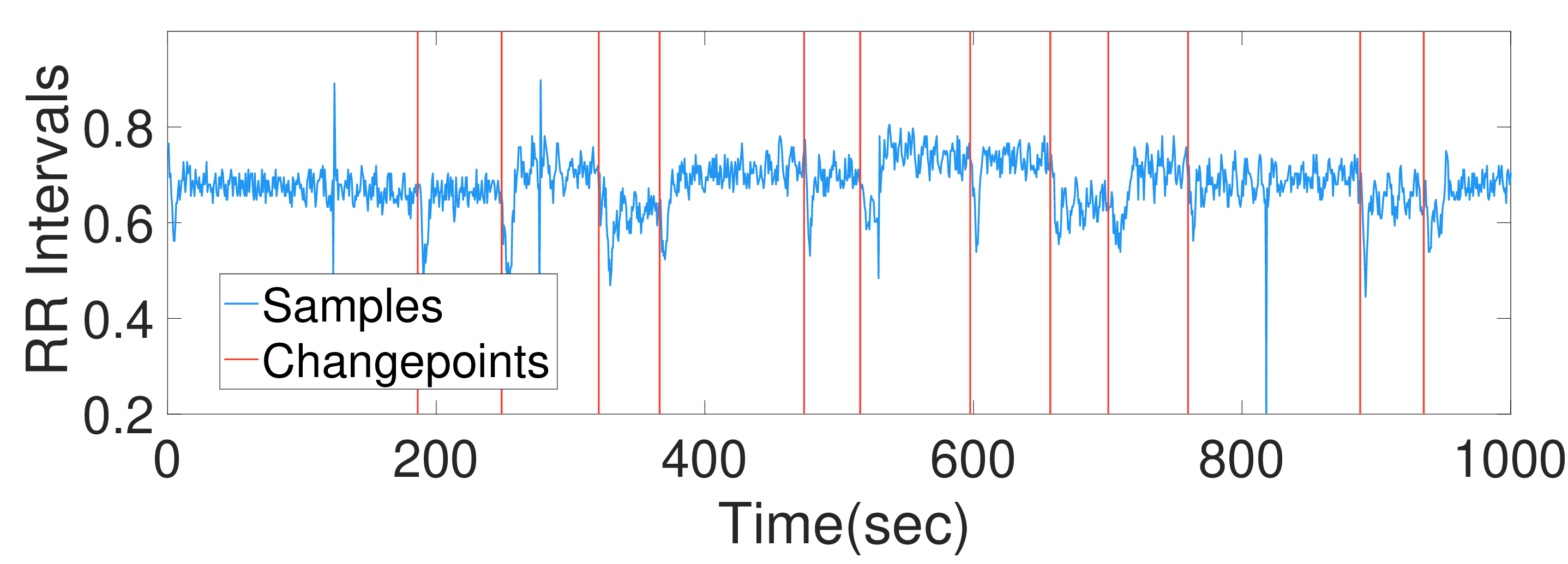}
    \caption{Artificial RR Interval data}
    \label{fig:rrgen_actgen}
    \vspace{-4.0mm}
\end{figure}

\subsection{Real data}
Real-world data used in this article came from the Physionet Cyclic Alternating Pattern Database \cite{terzano2002atlas}. ECG signals and hypnograms of patients with RBD and healthy controls were used.
Prior to analysis, the start times of ECG signals were adjusted to match the start of hypnograms. RR intervals were extracted from ECG signals and artifacts were removed using an open source validated HRV toolbox \cite{vest2017benchmarking}. Subject n16 did not have ECG signal and was removed from analysis. Data from 22 RBD patients and 15 controls were used in this study. 

\subsection{Segmentation methods}

\subsubsection{Recursive Mean Difference Maximization}
RMDM was proposed by Bernaola-Galv\`an \cite{Bernaola-Galvan2001} to study scaling behavior of the human heart rate, and has been tested on artificially generated non-stationary time series with different statistical properties and real data \cite{Bernaola-Galvan2012, fukuda2004heuristic}. The method recursively maximizes the difference in the mean values between adjacent segments. Given the input signal $S=\{x_1, x_2,...,x_N\}$ of length $N$, a sliding pointer is moved from the left to right, splitting the signal into $S_1=\{x_1, x_2,...,x_j\}$ and $S_2=\{x_{j+1},x_{j+2},...,x_N\}$ subsequences with $N_1$ and $N_2$ number of samples respectively, where $j$ is index at which the split occurs. Means of the subsequences $S_1$ and $S_2$ are calculated as:
\begin{equation} \label{eq:hrseg_1}
    \mu_1=\frac{1}{N_1}\sum_{x_i \in S_1}x_i, \,\,\,\,\, \mu_2=\frac{1}{N_2}\sum_{x_i \in S_2}x_i.
\end{equation}
The means are compared using the Student's $t$-statistic:
\begin{equation} \label{eq:hrseg_2}
    t(S_1,S_2)=\left|\frac{(\mu_{1}-\mu_{2})}{\sqrt{\sigma_P}}\right|
\end{equation}
where $\sigma_P$ is the pooled variance, defined as
\begin{equation} \label{eq:hrseg_3}
    \sigma_P=\frac{(N_1 + N_2)(V(S_1)+V(S_2))}{((N_1 + N_2 - 2) N_1 N_2}
\end{equation}
and $V(S)$ is the sum of squared deviations of the data in the signal $S$:
\begin{equation} \label{eq:hrseg_4}
    V(S)=\sum_{x_i \in S}(x_i-\mu)^2
\end{equation}

Student's $t$-statistic is calculated as a function of the index $j$ in the time series. A candidate changepoint $j_{max}$ is selected, at which $t(j)$ reaches the maximum $t_{max}$.

The significance level $\mathcal{P}(\tau)$ of $j_{max}$ is calculated as $\mathcal{P}(\tau)=\left\{t_{max}\leq\tau\right\}$, where $\mathcal{P}(\tau)$ could not be obtained in a closed analytical form and was numerically approximated by Bernaola-Galv\`an as
\begin{equation} \label{eq:hrseg_5}
    \mathcal{P}(\tau)=\left\{1-I_{\left[\frac{\nu}{\nu+\tau^2}\right]}(\delta\nu, \delta)\right\}^{\gamma}
\end{equation}
where $\gamma=4.19 \ln N - 11.54$, $\delta=0.40$, $N$ is the length of the signal, $\nu=N-1$ is the number of degrees of freedom, and $I_x(a,b)$ is the incomplete beta function.

If $\mathcal{P}(\tau)$ exceeds a predefined threshold $P_0$ ($0.95$ as in this work), the signal is split into two subsequences. Before index $j$ is confirmed as a changepoint, $\tau$ between the two candidate subsequences is also calculated to see if the significance exceeds $P_0$. To reduce false positives, the condition that the minimum candidate segment length should be greater than $l_0$ was added. This procedure is repeated for each new subsequence until splitting the signal into candidate subsequences that differ by a significance level $P_0$ is not possible.

\subsubsection{Binary Segmentation}
Binary segmentation (BiS) has been widely used in changepoint detection analysis and is computationally fast (\cite{chen2011parametric,killick2012optimal}). The procedure minimizes a cost function, and starts by searching for a changepoint $\tau$ in the input signal $S=\{x_1, x_2,...,x_N\}$ that satisfies the condition
\begin{equation} \label{eq:bs_1}
	C_{S_{1:\tau}} + C_{S_{(\tau+1):N}} + \beta < C_{S_{1:N}}
\end{equation}
where $C$ is a cost function and $\beta$ is a penalty term that reduces over-fitting. If the condition in \eqref{eq:bs_1} is met, $\tau$ becomes the first estimated changepoint, and $S_{1:\tau}$ and $S_{(\tau+1):n}$ become the first subsequences. The process continues within these segments until data cannot be divided any further. Any cost function $C$ can be adapted for use in this framework, and one or more changepoints can be detected. Here a BiS implementation from the Numerical Algorithms Group (NAG) Toolbox was used \cite{NAGtoolbox}.

\subsubsection{Pruned Exact Linear Time}
This algorithm was proposed by Killick {\em et al.} \cite{killick2012optimal} and minimizes a cost function which chosen according to prior probability distribution of data. Algorithm calculates following minimization
\begin{equation} \label{eq:pelt_1}
\begin{aligned}
	&F(s) = \min_{\tau\epsilon\tau_s}(\sum_{i=1}^{m+1}[C(S_{(\tau_{i-1}+1):\tau_i)}+\beta]) \\
    &= \min_{t}(\min_{\tau\epsilon\tau_t}(\sum_{i=1}^{m+1}[C(S_{(\tau_{i-1}+1):\tau_i)}+\beta])+C(S_{(t+1):n)}+\beta) \\
    &= \min_{t}(F(t)+C(S_{(t+1):n)}+\beta)
\end{aligned}
\end{equation}
PELT1 assumes that for all $t<m<T$, there is a constant K that satisfies
\begin{equation} \label{eq:pelt_2}
	C(S_{(t+1):m})+C(S_{(m+1):T})+K \leq C(S_{(t+1):T})
\end{equation}

After each changepoint is estimated, pruning is performed by removing points that satisfy the condition

\begin{equation} \label{eq:pelt_3}
	F(t)+C(S_{(t+1):m})+K \geq {F(m)}
\end{equation}
because these removed points cannot be the last optimal changepoint for $T>m$. In this work, NAG Toolbox implementation (``PELT1'') \cite{NAGtoolbox} and the built-in MATLAB r2016a function '\texttt{findchangepts.m}' (``PELT2'') was used. 

\subsubsection{Bayesian Blocks}

This algorithm was proposed by Scargle {\em et al.} and detects changepoints via dynamic programming \cite{Scargle2013}. The Bayesian Blocks (BBlocks) approach fits a piecewise constant signal model to data by maximizing a fitness measure specified according to the data type.

The total fitness of the partition $\mathcal{P}$ of the signal $S=\{x_1, x_2,...,x_N\}$ of length $N$ is additive and defined as
\begin{equation} \label{eq:bblocks_1}
    F\left[\mathcal{P}(S)\right]=\sum_{k=1}^{N_{s}}f(S_k)
\end{equation}
where $N_{s}$ is the number of segments and $f(S_k)$ is the fitness of the $k$th segment $S_k$ derived from maximizing the log likelihood of blocks given $M$ point measurements $x_{i}$,  $i \in 1, ..., M$
\begin{equation} \label{eq:bblocks_2}
    f(S_k) = \frac{(\sum_{M}x_i)^2}{4\sum_{M}^{}\sigma_i^2}
\end{equation}
where $x_{i}$ is the $i^\text{th}$ data point and $\sigma_{i}$ is the error variance of the data point measurement. Note that in our case, since a data point's measurement error was unknown and the signal was normalized, the error variance was taken as $\sigma=1$.

The BBlocks algorithm starts with a sub-signal of only one data point $x_{i=1}$, wherein only one segmentation is possible. In each step, a new datum $x_{i=i+1}$ is added to the signal and can be considered the last point of the last segment of a possible optimal segmentation of samples. The starting point $r$ of the last segment of this optimal partition is obtained at each step by the following
\begin{equation} \label{eq:bblocks_3}
   r_i=argmax[f(r)+F\left[\mathcal{P}^{opt}(r-1)\right]]
\end{equation}
where the definition of variables is the same as for the equations \eqref{eq:bblocks_1} and \eqref{eq:bblocks_2} and $\mathcal{P}^{opt}(r-1)$ is the optimal partition from the previous step.

When the last data sample is presented to the algorithm, the calculated value of $r_N$ becomes the last changepoint, marking the beginning of the last segment. This segment is removed, the data point before $r_N$ is considered the last datum, and the corresponding value of $r_i$ is assigned to the next changepoint. All changepoints are found by starting from the end, moving towards the start of the time series, and iteratively peeling off blocks.

\subsubsection{Bayesian Analysis of Changepoints}
BCP was proposed by Barry and Hartigan and implemented in the \texttt{R} programming environment by Erdman and Emerson \cite{barry1993bayesian,erdman2007bcp}. In this Bayesian approach, every observation has a probability density dependent on a parameter $\theta_{i}$ which is constant for all observations in the same segment. The segment containing observations ${x_{i},...,x_{j}}$ is denoted as ${x_{ij}}$. BCP is a product partition model in which the probability of change at each point is $p$ and the prior distribution of the parameter $\mu_{j}$ of a segment with data ${x_{ij}}$ is defined as $N(\mu_{0},\sigma_{0}^{2}/(j-i)$. After priors are set on parameters, the density of observations is calculated as 
\begin{equation} \label{eq:bcp_1}
	f_{ij}(x_{ij}) = \int{f_{ij}(x_{ij}|\theta)f_{ij}(\theta)d\theta}
\end{equation}
where $f_{ij}(\theta)$ is the prior distribution of parameters. Given a partition $\rho$ with changepoint locations $i(1)...i(b)$, the likelihood of observed data following this partition is given by
\begin{equation} \label{eq:bcp_2}
	\begin{aligned}
		L(x,\rho) &= K\prod_{r=1}^{b}{f_{i_{r-1}i_{r}}(x)c_{i_{r-1}i_{r}}} \\
    	L(x) &= \sum_{\rho}L(x,\rho)\\
    \end{aligned}
\end{equation}
where $c_{i_{r-1}i_{r}}$ is defined as ``prior cohesion''. Cohesion is the prior probability of observing a segment of given length.

A normal errors model is used, in which observations $x_{i}$ are assumed to follow a probability distribution $N(\mu_{i},\sigma^{2})$ and inter-arrival times are identically and geometrically distributed. Independent priors are specified for each of the parameters $p$ (the probability of change), $\mu_{0}$, $\sigma^{2}$, and $w = \sigma^{2}/(\sigma_{0}^{2} + \sigma^{2})$ which relates to the magnitude of change. The posterior probability of $\mu_{i}$ is estimated via a Markov sampling technique.

\subsubsection{Bayesian Online Changepoint Detection}

Adams and MacKay proposed a Bayesian Online Change point Detection (BOCD) algorithm \cite{adams-mackay-2007} and further developed by Turner {\em et al.} \cite{Turner2009}. This method estimates changepoints using Bayesian inference, whereby the posterior probability of the time since the last changepoint, referred to as ``run length'', is calculated sequentially.

For normally distributed data with unknown mean and variance, the conjugate prior on observations follows a normal-inverse-gamma distribution with $\nu$, $\alpha$, and $\beta$ hyper parameters. On each step of the algorithm, a new datum $x_{i=i+1}$ is added to the analyzed signal and $x_{t}$ indicates data sample at time t. The posterior predictive probability of a new datum has the form of a non-standardized Student's t-distribution \cite{Murphy} with $2\alpha$ degrees of freedom, center at $\mu$, and PPV $\frac{\alpha \nu}{\beta(\nu+1)}$.

The posterior predictive probability of segment length $r$ at data point $i$ is given by
\begin{equation} \label{eq:bocpd_1}
    \pi_t^{(r)} = t_{2\alpha_t}(x_t|\mu_t,\frac{\beta_t(\nu_t+1)}{\alpha_t \nu_t})
\end{equation}

Using predictive probability, growth probabilities are calculated as
\begin{equation} \label{eq:bocpd_2}
 \begin{aligned}
  P(r_t=r_{t-1}+1,x_{1:t}) = P(r_{t-1},x_{1:t-1})\pi_t^{(r)}(1-H(r_{t-1})
 \end{aligned}
\end{equation}
where $H(\tau)$ denotes the hazard function of a changepoint occurring. If intervals between changepoints are assumed to follow an exponential distribution with timescale $\lambda$, the hazard function becomes $H(\tau) = 1/\lambda$. The changepoint probability at time $t$ is
\begin{equation} \label{eq:bocpd_3}
 \begin{aligned}
  P(r_t=0,x_{1:t})=\sum_{r_{t-1}}P(r_{t-1},x_{1:t-1})P(x_t|r_{t-1},x_t^{(r)})H(r_{t-1})
 \end{aligned}
\end{equation}

The distribution of run lengths is calculated as
\begin{equation} \label{eq:bocpd_4}
 \begin{aligned}
  P(r_t|x_{1:t}) = \frac{P(r_t,x_{1:t})}{\sum_{r_t}P(r_t,x_{1:t})}.
 \end{aligned}
\end{equation}

Finally, the hyperparameters are updated according to
\begin{equation} \label{eq:bocpd_5}
  \begin{aligned}
    \mu_{t+1}&=\frac{\nu_t\mu_t + x_t}{\nu_t+1},  \\
    \nu_{t+1}&=\nu_t + 1,  \\
    \alpha_{t+1}&=\alpha_t + 0.5,  \\
    \beta_{t+1}&=\beta_t + \frac{\nu_t(x_t-\mu_i)^2}{2(\nu_t+1)} \\
  \end{aligned}
\end{equation}
as defined in the conjugate Bayesian analysis of the Gaussian distribution \cite{Murphy}. This process repeats for the remaining data samples of the analyzed signal. After the run length distribution is calculated for all samples, indices with maximum probabilities are used to estimate the location of changepoints.

\subsubsection{Modified BOCD Algorithm}
Each new datum added to the analyzed signal results in one of two possible events for the segment length $r_{t}$: 1) it increases so $r_{t} = r_{t-1} + 1$, or 2) a changepoint occurs and $r_{t} = 0$.

Originally, for each value of $t$, the run length probability vector $v$ is sequentially calculated. $v(1) = P(r_{t}=0)$ is calculated by \eqref{eq:bocpd_3}. $v(2:t)$ is the probability of each possible run length at time $t$ and calculated by \eqref{eq:bocpd_2}. This vector is normalized by \eqref{eq:bocpd_4}. Finally, all run length probability vectors are concatenated, and the run length with the maximum probability is selected as the run length for time $t$.

\begin{figure}[htbp]
    \centering
    \includegraphics[width=0.95\linewidth]{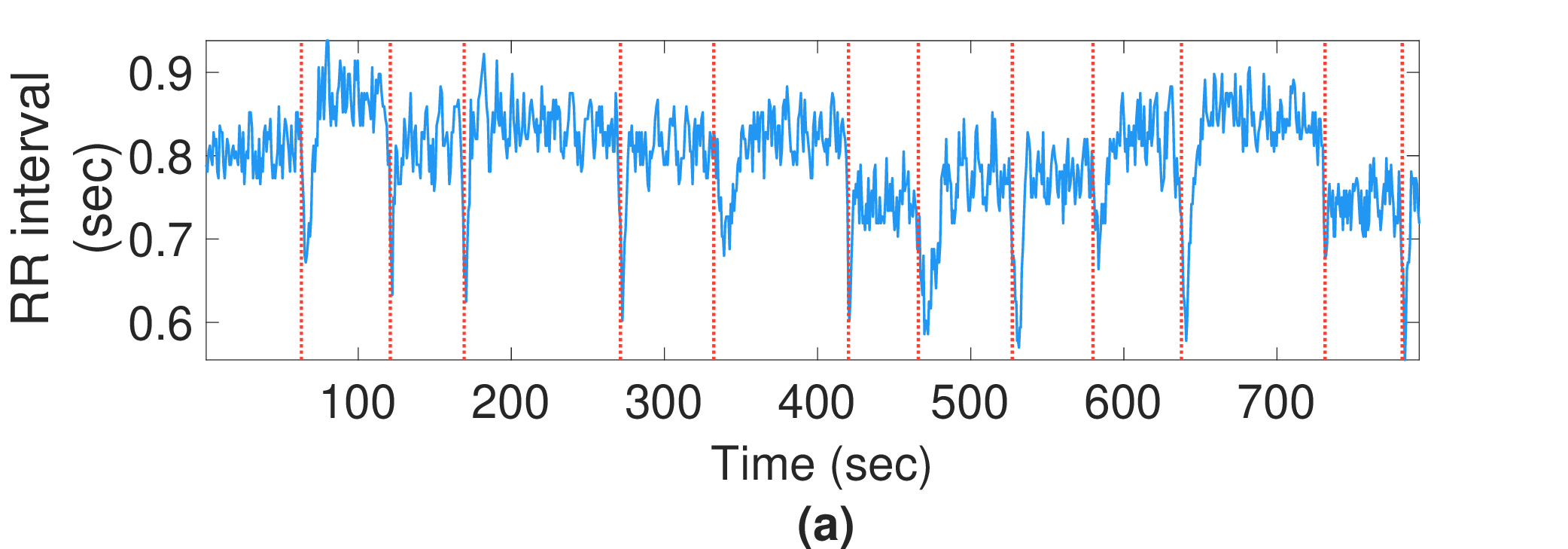}
    \includegraphics[width=0.95\linewidth]{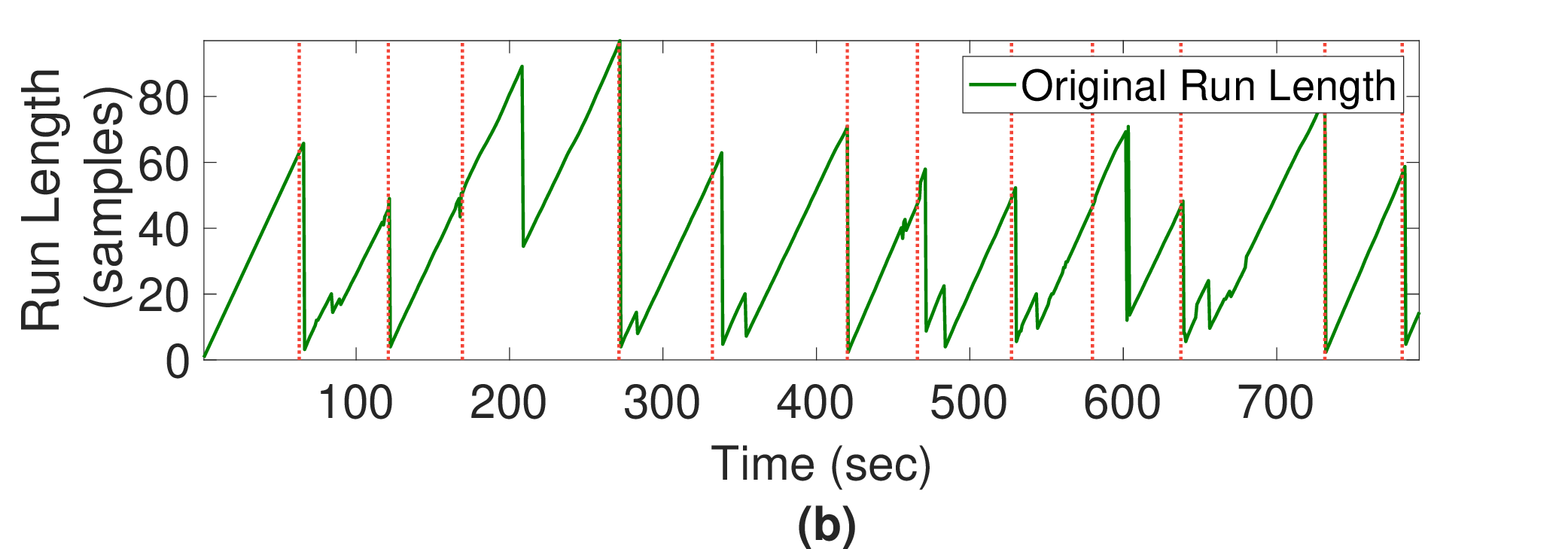}
    \includegraphics[width=0.95\linewidth]{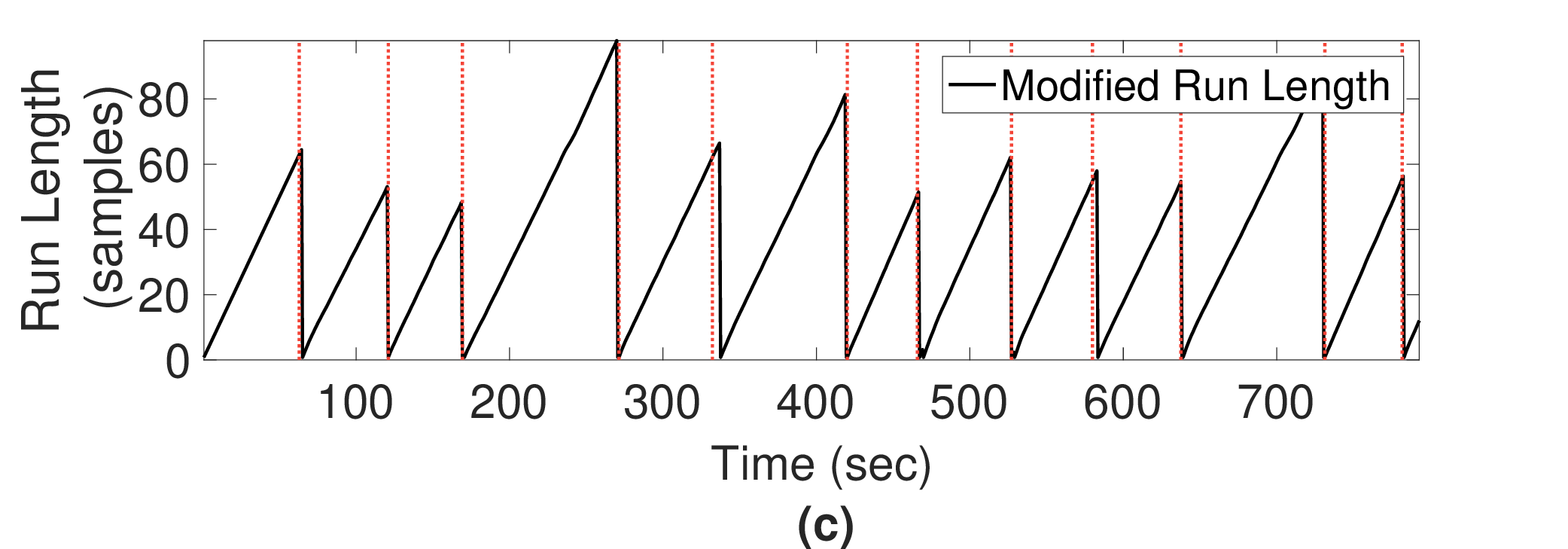}
    \caption{(a) Artificially generated RR interval data, true changepoints are shown in red lines. (b) Run length for BOCD. (c) Run length for mBOCD. Note that the false triggering of BOCD algorithm is avoided using the mBOCD algorithm.}
    \label{fig:bocpd1}
\end{figure}

Although the run length should drop to zero as a changepoint is encountered in the time series, this approach fails to find $r_{t} = 0$. Due to sequentially calculating run length probability vectors, $v(1)$ is rarely the maximum row in the vector, so the run length is almost never set to zero. This effect can be seen in the middle plot of Figure \ref{fig:bocpd1}, which illustrates the performance of the original BOCD algorithm.

A simple modification enables the correct selection of $r_{t} = 0$ when changepoints are encountered. The growth probability, e.g. probability of continuing the current run, is calculated as
\begin{equation} \label{eq:bocpd_7}
 \begin{split}
  P(r_t=r_{t-1}+1|x_{1:t})=& \sum_{r_{i-1}}P(r_{t-1}|x_{1:t-1})\\&\times P(x_t|r_{t-1},x_t^{(r)})(1-H(r_{t-1}))
  \end{split}
\end{equation}
\vspace{-5.0mm}

The growth probability is compared to the changepoint probability $P(r_t=0|x_{1:t})$. If the changepoint probability is higher, the segment is ended and the run length becomes zero. Otherwise, the run length vector is increased by one. This approach is illustrated in Figure \ref{fig:bocpd1} and is denoted as the modified (mBOCD) algorithm.

\subsection{Performance metrics for changepoint detection}
In order to assess how well estimated changepoints mapped to true changepoints, the following performance measures were used:
\begin{itemize}
\item\textbf{True Positive (TP)}: An estimated changepoint within a temporal tolerance $\gamma$ of a true changepoint was labeled as a true positive. If more than one estimated changepoint occurred within $\gamma$ of a true changepoint, only one estimated changepoint was counted towards the total number of true positives. 
\item\textbf{False Positive (FP)}: If the estimated changepoint was not within the tolerance, $\gamma$, of any true changepoint, then it was labeled as a false positive. 
\item\textbf{False Negative (FN)}: If there was not any estimated changepoint within the tolerance $\gamma$ of a true changepoint, then a false negative was recorded.  
\end{itemize}
True negatives cannot be counted, since these would overwhelm any statistics and depend heavily on the sampling frequency. The recall or true positive rate (${\displaystyle {\mathit {TPR}}={\mathit {TP}}/({\mathit {TP}}+{\mathit {FN}})}$) and positive predictive value (${\mathit {PPV}} = {\mathit {TP}}/({\mathit {TP}}+{\mathit {FP}})$) were calculated in this analysis. The true negative rate was not calculated because there were many more non-changepoints versus changepoints in the data. Instead, the number of false positives was counted. Finally, the F1 score –- the harmonic mean of TPR and PPV (${\mathit {F1}}=2{\mathit {TP}}/(2{\mathit {TP}}+{\mathit {FP}}+{\mathit {FN}})$) -  was used to optimize the parameters of each algorithm.

\subsection{Parameter selection}

To find optimal parameters for each algorithm, the F1 score was maximized via grid search using realistic search ranges for each parameter. For RMDM, the minimum segment length ranged from $l_0=\left\{6:1:12\right\}$. For BBlocks, the free parameter ranged from $\gamma=\left\{1.5:0.5:4,5\right\}$. For BCP, parameters $w_{0}$ and $p_{0}$ ranged from $\left\{0:0.1:1\right\}$. After $w_{0}$ and $p_{0}$ was set, the optimal posterior probability cut-off level was searched for in the range $\left\{0.5:0.1:1\right\}$. If the posterior probability of a point was higher than that level, it was labeled as a changepoint. In the analysis using the BOCD method, the expected segment length was tested in the range $\lambda=\left\{100:100:2000\right\}$. For mBOCD, the expected segment length ranged from $\lambda=\left\{10:10:100\right\}$ for RRGen.

PELT2 was tested via detecting changes in root mean square level, standard deviation, mean, and ``linear" mode -- which finds the locations at which the mean and the slope of the signal change most abruptly. For PELT1 and BiS, changes in mean or mean and standard deviation were evaluated at the same time, and assumed data were drawn from normal distributions. For Hannan-Quinn, BIC, and AIC information criteria, the following penalties were tested respectively: $\beta = 2p\times\log(\log(N))$, $\beta = p\times\log(N)$, and $\beta = 2\times p$.

To analyze artificial RR interval data, for RMDM a minimum segment length $l_0=7$ was used. For BiS changes based on mean were evaluated with a Hannan-Quinn penalty. For PELT1 changes based on mean were evaluated with a BIC penalty, while for PELT2 changes were based on RMS. For Bayesian Blocks a free parameter value $\gamma=4$ was used. For BCP values of $w_{0}=0.2$, $p_{0}=0.3$, and a posterior probability cut-off level of 0.6 were used. For BOCD, $\lambda=1840$ was used and for mBOCD an expected segment length $\lambda=80$ was used.

\subsection{Classification}
\label{subsec:Classification}
In this section, a novel approach to classifying patients is presented, and used to evaluate the utility of each changepoint detection method. In this proposed method, the lengths of time between estimated changepoints was fitted to a Pareto distribution which is characterized by scale and shape parameters. These two parameters were extracted for each subject and used as features.  For classification, a simple K-Nearest Neighbors (KNN) approach with 10-fold cross validation was used. In order to find the optimal number of neighbors and distance metric in KNN, Bayesian optimization was performed. The objective function of KNN is defined as the percentage of neighbors belonging to the same class for each point, and the highest area under the precision-recall curve (AUCPR) was calculated using this metric. We note that KNN was chosen as an example to illustrate the technique, rather than an optimal classifier. Performance metrics calculated were Accuracy (ACC), Area Under the Receiver Operating Characteristic Curve (AUROC), AUCPR, TPR, PPV and F1 score.

\begin{table}[h]
\caption{Performance of changepoint detection algorithms using RRGen data, evaluated on 2800 hours of artificial data (400 different tachograms). Values shown are mean $\pm$ one standard deviation.} 
\centering 
\begin{tabular}{c c c c} 
\hline  \\[0.5ex] 
Methods & TPR & PPV & \multicolumn{1}{p{1.6cm}}{\centering Number of\\ FP/Hour} \\ [1ex] 
\hline\hline \\ [0.5ex] 
RMDM    & \textbf{0.91 $\pm$ 0.02} & 0.44 $\pm$ 0.01 & 41 $\pm$ 2 \\
BiS     & 0.79 $\pm$ 0.04 & 0.49 $\pm$ 0.01 & 28 $\pm$ 2 \\
PELT1   & 0.74 $\pm$ 0.05 & 0.54 $\pm$ 0.01 & 21 $\pm$ 2  \\
PELT2   & 0.90 $\pm$ 0.02 & 0.50 $\pm$ 0.01 & 30 $\pm$ 1  \\
BBlocks & 0.80 $\pm$ 0.03 & 0.52 $\pm$ 0.01 & 25 $\pm$ 2  \\
BCP     & 0.80 $\pm$ 0.03 & 0.50 $\pm$ 0.02 & 27 $\pm$ 2  \\
BOCD    & 0.78 $\pm$ 0.02 & 0.31 $\pm$ 0.01 & 61 $\pm$ 4 \\
mBOCD   & 0.87 $\pm$ 0.02 & \textbf{0.84 $\pm$ 0.02} & \textbf{6 $\pm$ 1}  \\ [1ex]
\hline 
\end{tabular}
\label{table:rrgenPerf} 
\end{table}

\section{Results}

\subsection{Changepoint detection performance on artificial RR interval data}

While RMDM achieved highest TPR when applied to artificial RR interval data, mBOCD had higher PPV and fewer false positive counts compared to other methods (see Table \ref{table:rrgenPerf}). When adding artificial noise (Probability of noise = $0:0.005:0.015$), RMDM was found have highest TPR compared to other methods in all levels of noise. Overall PPV was highest for mBOCD, but this metric declined steeply with the addition of noise. The addition of ectopic beats (probability of ectopic beat = $0:0.005:0.015$) minimally affected TPR for all changepoint algorithms. PELT2 was most robust to ectopic beast in terms of both TPR and PPV. Performance as a function of tolerance (Detection tolerance = 1:2:7) was also evaluated. TPR of BOCD method was affected more than other algorithms when tolerance was increased. mBOCD had the highest PPV among all methods for all tolerance values.

\subsection{Evaluation of changepoint detection on real data}

Changepoint detection on real data was evaluated by using the distributions of the intervals between changepoints (fit to a Pareto distribution) as explained in section \ref{subsec:Classification}. The scale and shape parameters of the Pareto distribution were used as inputs for KNN classifier.  Table \ref{table:classify} shows that BiS achieved highest AUCPR (0.91) for identifying RBD from the distribution of intervals between detected changepoints, and RMRD the highest AUROC and accuracy, both at 0.89.

\begin{table}[h]
\caption{Classification of normal and REM Behavior Disorder patients} 
\centering 
\begin{tabular}{c c c c c c c} 
\hline  \\[0.5ex] 
Methods & ACC & AUCROC & AUCPR & TPR & PPV & F1 \\ [1ex] 
\hline\hline \\ [0.5ex] 
RMDM    & {\bf 0.89} & {\bf 0.89} & 0.88 & 0.87 & 0.87 & 0.87 \\
BiS     & 0.86 & 0.88 & {\bf 0.91} & 0.81 & 0.87 & 0.84 \\
PELT1   & 0.86 & 0.80 & 0.83 & 0.86 & 0.80 & 0.83 \\
PELT2   & 0.86 & 0.79 & 0.84 & 0.86 & 0.80 & 0.83 \\
BBlocks & 0.78 & 0.77 & 0.69 & 0.73 & 0.73 & 0.73 \\
BCP     & 0.68 & 0.67 & 0.62 & 0.59 & 0.67 & 0.63 \\
BOCD    & 0.73 & 0.72 & 0.66 & 0.67 & 0.67 & 0.67 \\
mBOCD   & 0.76 & 0.75 & 0.73 & 0.69 & 0.73 & 0.71 \\ [1ex]
\hline 
\end{tabular}
\label{table:classify} 
\vspace{-6.0mm}
\end{table}

\section{Discussion}
In this work, the performance of six commonly available state-of-the-art changepoint detection approaches, with two additional modified algorithms, were extensively compared in cardiovascular time series data. One modified algorithm was presented here for the first time as an improvement to the newest of all presented algorithms, a modified BOCD approach. Finally, a novel approach for classifying patients based on distributions of changes points was presented and tested as a function of chosen changepoint detection algorithm. Two different artificially generated datasets were used to evaluate the generalizability of each algorithm and to optimize parameters on known changepoints. Changepoint detection accuracy and sensitivity to noise, ectopy, and temporal tolerance varied by algorithm. The mBOCD algorithm demonstrated the highest PPV on artificially generated RR interval data, whereas PELT2 was the most robust to ectopic beats. 

All algorithms were then utilized in a classification task. Improvements in downstream classification performance would indicate that the segmentation was successful since the algorithm was able to detect alterations due to disease. Distribution parameters were extracted from changepoints and a KNN classifier was trained to classify RDB patients from controls using these parameters as features. Parameters based on the RMDM-based changepoints achieved an TPR of 0.87 and accuracy of 0.89 much higher than an actigraphy based classification approach reported in the literature, which reported a specificity of 0.96, but at with a very low sensitivity of 0.20 and an equivalent accuracy of 0.58 for a balanced dataset \cite{louter2014actigraphy}. 
This result could be due to the fact that actigraphy based methods over-estimate wake epochs \cite{quante2018actigraphy}. Moreover, RBD patients do not have dream enactment episodes every night so actigraphy signal might not have artifacts of disease constantly. However, since heart rate changes often are an insensitive marker of movement and likely carry less information, it is more likely that the approach outline in this paper provided a more accurate approach to quantifying changes in activity, since activity metrics are essentially very coarse summaries of the overall changes exhibited by a subject.  (For example, typical activity metrics include L5, the average activity during the least active 5-h period, and M10, the average activity during the most active 10-h period.)

\section{Conclusion}
CPD should be optimized for a given application and classifier. If the detected changepoints are correct, changepoints should reflect the structure altered by physiological condition and a classifier can be designed using the changepoints. Once optimized, it is possible to use CPD to provide an accurate classifier for diseases where differences in the statistics of movement or state changes are likely to be observed. All algorithms and data used in this work are available online \cite{cakmak2018cpdtoolbox} or, where licensing restricts the distribution, can be downloaded from links in the toolbox to the algorithms. 

\section{Acknowledgments}
The authors wish to acknowledge the support of the National Science Foundation Award 1636933, National Institutes of Health (Grants  HL127251, P50 HL117929, K01ES025445, and R01HL136205) and the Rett Syndrome Research Trust. Any opinions, findings, and conclusions or recommendations expressed in this material are those of the author(s) and do not necessarily reflect the views of the sponsors.

\bibliographystyle{unsrt}
\bibliography{references}
\end{document}